\pdfoutput=1

\documentclass[11pt]{article}

\usepackage{acl}
\usepackage{times}
\usepackage{latexsym}

\usepackage{soul}
\usepackage{times}
\usepackage{empheq}
\usepackage{latexsym}
\usepackage{booktabs}
\usepackage{multirow}
\usepackage{subfigure}
\usepackage{tabularx}
\usepackage{color}
\usepackage{graphicx}
\usepackage{caption}

\usepackage{float}
\usepackage{rotating}

\usepackage[T1]{fontenc}

\usepackage[utf8]{inputenc}

\usepackage{microtype}

\usepackage{inconsolata}

\title{Rethinking Transformer-based Multi-document Summarization: An Empirical Investigation}

\author{
    Congbo Ma$^{1}$, Wei Emma Zhang$^{2}$, Dileepa Pitawela$^{2}$, Haojie Zhuang$^{2}$, Yanfeng Shu$^{3}$\\
    $^{1}$Macquarie University, Sydney, Australia, 
    $^{2}$The University of Adelaide, Adelaide, Australia, \\
    $^{3}$CSIRO, Australia. 
    \texttt{congbo.ma@mq.edu.au}, \texttt{yanfeng.shu@data61.csiro.au}\\
    \texttt{\{wei.e.zhang, dileepa.pitawela, haojie.zhuang\}@adelaide.edu.au}, 
     \\
}

\begin{document}
\maketitle
\begin{abstract}
The utilization of Transformer-based models prospers the growth of multi-document summarization (MDS). Given the huge impact and widespread adoption of Transformer-based models in various natural language processing tasks, investigating their performance and behaviors in the context of MDS becomes crucial for advancing the field and enhancing the quality of summary. To thoroughly examine the behaviours of Transformer-based MDS models, this paper presents five empirical studies on (1) measuring the impact of document boundary separators quantitatively; (2) exploring the effectiveness of different mainstream Transformer structures; (3) examining the sensitivity of the encoder and decoder; (4) discussing different training strategies; and (5) discovering the repetition in a summary generation. The experimental results on prevalent MDS datasets and eleven evaluation metrics show the influence of document boundary separators, the granularity of different level features and different model training strategies. The results also reveal that the decoder exhibits greater sensitivity to noises compared to the encoder. This underscores the important role played by the decoder, suggesting a potential direction for future research in MDS. Furthermore, the experimental results indicate that the repetition problem in the generated summaries has correlations with the high uncertainty scores.
\end{abstract}

\section{Introduction}\label{sec1}

The innovation and contemporary developments of Transformer architecture \cite{Attention-is-all-uneed} thrives multi-document summarization (MDS) \cite{ma2020multi}. This motivates us to study the behaviors of the Transformer structure MDS models. Through these analyses, we aim to provide a thorough understanding of MDS and its intricacies within the MDS model framework. We undertake a comprehensive investigation from five distinct perspectives covering the Transformer-based MDS model design pipeline:
(1) Document input perspective: we conduct experiments to quantitatively assess the impact of document boundary separators from a standpoint of document input; 
(2) Transformer structure perspective: we explore the effectiveness of different mainstream Transformer structures; 
(3) The significance of encoder and decoder perspective: we design empirical studies by adding noises on top of the encoder and decoder; 
(4) Training strategy perspective: we restructure the source documents and include self-supervised learning; 
(5) Summary generation perspective, we explore the uncertainties when repetition problems occur in the summary generation process. 

The primary distinction between SDS and MDS lies in the variance of source document numbers. One straightforward way that convert MDS to SDS is concatenating text spans and processing them as a flat sequence \cite{liu2018generating, chu2019meansum, Brazinskas2020few, mao2020multi, Chao2022Read}. One way to aid the models in detecting and modeling document-to-document relationships in one flat sequence is to utilize document boundary separators \cite{Alexander2019multinews, wen2022PRIMERA}. However, there is a notable gap in the current literature regarding a qualitative and quantitative examination of the influence of document boundary separators. This absence of exploration serves as the driving force behind our initiative to investigate whether these separators contribute to enhanced model performance and foster awareness of document boundaries within the feature space of MDS models. Through experiments conducted on three distinct Transformer structures, we discerned that the impact of document boundary separators varies among models with differing hierarchies.
Uncertainty analysis is a pivotal approach employed in the examination and assessment of generation systems \cite{xu2020understanding} which can serve as an important indicator to show how the model performs during the summary generation. We then investigate the variation of summary prediction uncertainty by exploring the relations between separators and the predictive uncertainty of the structures. Certainly, measuring uncertainty in the context of summarization can provide insights into how the presence of document boundary separators affects the behavior of Transformer-based models and their summarization outcomes. By quantifying uncertainty through the entropy calculations, we gain a deeper understanding of the level of confidence or ambiguity the model has in its generated summaries.

Instead of simply concatenating all the input documents into a flat sequence and applying SDS models, the hierarchical Transformer structure \cite{liu-lapata-2019-hierarchical, pasunuru2021efficiently, li2020leveraging} has been proposed to specifically solve MDS tasks. This structure has been used for encoding multiple documents in a hierarchical manner, enabling the capture of cross-document relations through the utilization of an attention mechanism. The hierarchical Transformer structure contains a low-level Transformer that encodes tokens and a high-level Transformer that is used to encode coarser-grained textual units. This motivates us to further explore the influence of different hierarchies on MDS performances. We explore the effect of different granularity of high-level Transformer on the performance of MDS models. In this paper, we consider sentence-level and document-level features as different granularities. Based on the empirical studies, our findings indicate that for MDS tasks involving relatively short documents, flat Transformer models are a suitable choice. Also, the hierarchical structure prefers higher granularity in high-level Transformer structures.

In addition to exploring the hierarchical structure of Transformer-based MDS models, we explore the Transformer's internal structure. Based on the existing Transformer-based MDS methods, we find that many of the MDS models focus on modifying the components of encoder \cite{liu-lapata-2019-hierarchical, pasunuru2021efficiently, liu2021highlight, ma2022document} and fewer works pay attention to ameliorating the decoder \cite{han2020abstractive, liu2022BRIO} to cater the requirements for MDS tasks. This motivates us to explore the robustness of the encoder and decoder towards interference under the same noise conditions. Therefore, we add Gaussian noises at the parameter space of the encoder or decoder to fulfill this purpose. The experimental results indicate that the decoder exhibits greater sensitivity compared to the encoder in MDS scenarios. This finding underscores the need for increased attention to decoder enhancements in future research within the MDS community.

Based on the analysis of Transformer-based MDS models, we also pay attention to exploring different training strategies for further enhancing the performance of MDS models. Different training strategies offer unique approaches to utilize available data and optimize model performance. By investigating diverse training strategies, we aim to identify the most effective methods for training MDS models, leveraging the characteristics of the dataset and the summarization task at hand. These strategies involve using pseudo datasets, fine-tuning on original datasets, or a combining of both. To generate pseudo data, we treat individual documents in a document set as pseudo-summaries and create multiple sets of pseudo-document-summary pairs. We evaluate three training approaches: training exclusively on the pseudo dataset, mixing the pseudo dataset with the original dataset, and a two-step process of training on the pseudo dataset followed by fine-tuning on the original dataset.
The experimental results demonstrate that the pretrain-finetune strategy consistently outperformed the other training strategies, leading to improved summarization quality. The analysis of feature distributions further supported this finding, highlighting the alignment between the finetuned model and the baseline model. These results provide valuable insights into the effectiveness of the pretrain-finetune approach in enhancing summarization performance. The findings of this study can guide future research and development in the field of abstractive summarization, emphasizing the importance of training strategies for achieving higher-quality summaries.

Moreover, while the different Transformer structures and training strategies demonstrated variations in performances, an observation is the presence of repetitive patterns in the generated summaries, indicating a potential issue that needs to be addressed in abstractive summarization systems. 
Liu et al. \cite{liu2023reducing} gave two possible reasons behind the repetition problem in abstractive summarization: (1) attending to the same location in the source and (2) attending to similar but different sentences in the source. 
In this paper, we explore the cause of repetitive problems in abstractive summarization by examining predictive uncertainty. We quantify uncertainty scores at each time slot during the summary generation process. The analysis aims to observe how the uncertainty score changes when repetition phenomena occur, allowing us to identify positions where uncertainty is localized in repetitive behavior. 
The analysis reveals that as the model generates repetitive sentences or words, the uncertainty score rises, pointing out decreased confidence and increased uncertainty regarding the appropriateness and relevance of repeated elements in the summary. Understanding this relationship allows us to develop strategies to mitigate repetition and improve the quality of generated summaries.

\section{Methodology}

We introduce how to design the MDS experiments from the following angles: input data, Transformer structures, training strategies and summary generation. Therefore, we design five experiments to evaluate the behaviors of Transformer-based MDS models: (1) the measurable impact of document boundary separators; (2) the effectiveness of different Transformer structures;
(3) the sensitivity of the encoder and decoder; (4) different training strategies; (5) repetition in document generation.

\subsection{The Measurable Impact of Document Separators}

We modify the source documents instead of the summarization models to the format of: $\mathcal{D} = \{\mathbf{d}^1, \mathbf{s}, \mathbf{d}^2, \mathbf{s}, ..., \mathbf{s}, \mathbf{d}^N \}$, where $N$ is the number of documents in a document set $\mathcal{D}$, the superscript $\mathbf{d}^n$ represents the n-th document in the set, and $\mathbf{s}$ denotes the special tokens. We investigate different Transformer models on two MDS datasets and eleven evaluation metrics to explore the impact of the document boundary separators qualitatively and quantitatively.
We analyze and compare the prediction uncertainty from different datasets and different formats of source documents by inspecting entropy values during summary generation. We aim to understand how decisions by adding document boundary separators are reflected in the model’s uncertainty.
In the generation process, each predictive position $\mathbf{X}_i$ has an outcome probabilistic distribution $\mathbf{x}_{i1},..., \mathbf{x}_{im} $, $m$ is the number of a corpus pool. We use entropy as an uncertainty measurement which can be calculated as follows:

\begin{equation} \small
\begin{aligned}
H(\mathbf{X}_i)=-\sum_{j=1}^{m}P(\mathbf{x}_{ij}) \ logP(\mathbf{x}_{ij})
\end{aligned}
\label{equ:entropy}
\end{equation}
Because the size of the corpus pool is large and the prediction distribution is usually long-tailed \cite{xu2020understanding}, we sort the prediction distribution $\mathbf{X}_i$ in descending order and get a minimal set of tokens where the sum prediction values are larger than 0.95, and then normalize the distribution. We calculate the entropy value based on the new distribution $P'(\mathbf{x}_{ij})$.
The utilization of entropy as a measure allows us to gauge the distribution of probabilities across different tokens within the predictive positions of the summaries. Higher entropy values indicate a wider spread of probabilities, suggesting that the model is less certain about the most appropriate token to choose. Conversely, lower entropy values suggest that the model is more confident in its token predictions. The quantification of uncertainty through entropy measurements and its qualitative analysis enables us to assess how the introduction of document boundary separators influences the performance of the summaries generated by Transformer-based models. This holistic approach helps us unravel the nuanced impact of document boundary separators on the MDS process and gain valuable insights into the behavior of these models in handling multiple document inputs.

\subsection{The Effectiveness of Different Transformer Structures}
\label{sec: transformer structures}

Transformer structures have become an essential component of many state-of-the-art natural language processing models. However, the design of the Transformer architecture can vary dramatically, and different structures may impact the performance of the model on different tasks. In this study, we aim to evaluate the effectiveness of different Transformer structures for MDS tasks. Specifically, we focus on two types of structures: flat Transformer and hierarchical Transformer. 

The flat Transformer consists of a single layer of self-attention and feed-forward neural network layers that process the input tokens sequentially. In contrast, the hierarchical Transformer has a more complex structure, where the input tokens are first grouped into sentences or documents, and then processed by local and global Transformer layers. To explore the hierarchical Transformer structure, we investigate two different granularities of high-level Transformer: sentence-level and document-level. Building on the work of Liu \cite{liu-lapata-2019-hierarchical}, we make modifications to the local Transformer layers to encode individual documents. The global Transformer layers are then able to exchange information at the sentence or document level.

Our analysis is motivated by the need to better understand how different Transformer structures can impact the performance of MDS models. By comparing the performance of the flat Transformer and hierarchical Transformer structures, we aim to identify which structure is more effective for multiple document summarization data.

\subsection{The Sensitivity of Encoder and Decoder}

\label{sec: encoder and decoder}

In summarization tasks, the encoder plays a crucial role in extracting representations from the input text, while the decoder is responsible for generating the output summary, which requires producing coherent and meaningful language. Given the intricate nature of summary generation, the decoder's role demands fine-grained control and precision, making it potentially more sensitive than the encoder.
To explore the sensitivity of the encoder-decoder in Transformer-based summarization models, we add Gaussian noise at the parameter space of the encoder or decoder. We devise this experiment based on the intuition that a module (whether it's the encoder or decoder) exhibits varying sensitivity to noise, thereby signifying the differing degrees of importance each module holds for overall performance. Formally, we have:

\begin{equation} \small
\begin{aligned}
z=f(x;\Theta + \alpha n), n\sim N(\mu, \delta)
\end{aligned}
\end{equation}

\noindent where $f(\cdot )$ is the component in Transformer; $\Theta$ is the parameters in $f(\cdot )$; $n$ represents Gaussian noise; $\mu$, $\delta$ are mean and variance in the Gaussian noise, $\alpha$ is the weighted factor.

\subsection{Different Training Strategies}
\label{sec: training strategies}
In this study, we aim to investigate the impact of different training strategies on Transformer models for abstractive summarization. While we have previously examined the components of Transformer models, the specific influence of training strategies remains unexplored. Our objective is to identify the most effective training strategies by leveraging the inherent characteristics of MDS datasets, without the need for external data sources. To create pseudo data utilizing the characteristic MDS, we adopt a straightforward approach. We treat one document from a given document set as a pseudo-summary while considering the remaining documents as input documents. This process is iterated, systematically selecting each document in the set as a pseudo-summary, until all input documents have served as pseudo-summaries. Consequently, we generate multiple sets of pseudo-document-summary pairs, which we refer to as pseudo-MDS dataset. The original MDS dataset is denoted as the original dataset in the subsequent analysis.

To evaluate the effectiveness of different training strategies, we design three distinct approaches. Firstly, we train the MDS model exclusively on the pseudo dataset. Secondly, we mix the pseudo dataset with the original dataset, creating a comprehensive mega dataset, on which the MDS model is trained. Lastly, we employ a two-step process, initially training the model on the pseudo dataset and subsequently fine-tuning it on the original dataset.

\subsection{Repetition in Document Generation}

For abstractive MDS, a persistent challenge arises from the inclination of models to produce repetitive sentences or words during the summarization process. This tendency creates a loop that is difficult to break, hampering the generation of accurate summaries. To analyze what may cause repetitive problems, we delve into an analysis of prediction uncertainty, examining uncertainty scores throughout the generation process and localizing uncertainty to certain positions in a repetition behavior. 

To quantify uncertainty, we employ Equation \ref{equ:entropy}, which calculates the uncertainty score for each time slot during the summarization generation. By applying this equation, we obtain a measure of uncertainty that corresponds to the level of doubt or ambiguity associated with the generated output. The analysis focuses on observing how the uncertainty score evolves in response to the occurrence of repetition phenomena.

\section{Empirical Studies and Analyses}

\subsection{Settings for Empirical Studies} 

We evaluate the performance of three Transformer models: Vanilla Transformer (VT) \cite{Attention-is-all-uneed}, Vanilla Transformer with copy mechanism (VTC), and modified Hierarchical Transformer (HT) \cite{liu-lapata-2019-hierarchical}. These models are assessed on two widely used MDS datasets: Multi-XScience \cite{Yao2020multi} and Multi-News \cite{Alexander2019multinews}. To comprehensively analyze their performance, we employ eleven evaluation metrics: ROUGE \cite{lin-2004-rouge} including ROUGE-1 (R-1), ROUGE-2(R-2), ROUGE-L (R-L), ROUGE-SU (R-SU), ROUGE-WE (R-WE) \cite{ng-abrecht-2015-better}, BLEU \cite{Papineni-etal}, S\textsuperscript{3} \cite{peyrard-etal-2017-learning} including pyramid (pyr) and responsiveness (resp) scores, BertScore (BS) \cite{Zhang2020BERTScore}, Relevance (Rel) \cite{Simple-Theoretical}, Redundancy(Red) \cite{Simple-Theoretical}.

%
%

\begin{table*}[htp] \small
\centering
\resizebox{0.9\textwidth}{!}{
\begin{tabular}{|c|c||c|c|c|c|c|c|c|c|c|c|}
\hline
Datasets & Models    & R-1$\uparrow$ & R-2$\uparrow$ & R-L$\uparrow$ & R-SU$\uparrow$& R-WE$\uparrow$ & BLEU$\uparrow$ & S\textsuperscript{3} (pyr/resp)$\uparrow$ & BS$\uparrow$ & Red$\downarrow$ & Rel$\uparrow$ \\ \hline \hline
\multirow{6}{*}{\begin{tabular}[c]{@{}c@{}}Multi\\ -XScience\end{tabular}} & VT        & 0.2714 & 0.0490 & 0.1030 & 0.0784 & 0.1523 & 2.9773  & 0.2103/0.3609 & 0.5330 & -4.0712 & -5.8352  \\  
                                                                           & VT w/o S  & 0.2670 & 0.0480 & 0.1553 & 0.0767 & 0.1580 & 3.3623  & 0.2202/0.3663 & 0.5405 & -6.1908 & -4.8609  \\ 
                                                                           & VTC       & 0.2635 & 0.0483 & 0.1499 & 0.0734 & 0.1659 & 4.6037  & 0.2561/0.3885 & 0.5590 & -7.0585 & -4.5802  \\  
                                                                           & VTC w/o S & 0.2713 & 0.0468 & 0.1502 & 0.0780 & 0.1702 & 4.7615  & 0.2554/0.3861 & 0.5621 & -7.8402 & -4.2908  \\ 
                                                                           & HT        & 0.2571 & 0.0483 & 0.1615 & 0.0692 & 0.1407 & 7.1501  & 0.1769/0.3473 & 0.5303 & -4.6987 & -8.0379  \\  
                                                                           & HT w/o S  & 0.2216 & 0.0376 & 0.1446 & 0.0521 & 0.1100 & 5.2862  & 0.1428/0.3295 & 0.5108 & -4.0142 & -11.6068 \\ \hline
\hline                                                                  \multirow{6}{*}{\begin{tabular}[c]{@{}c@{}}Multi\\ -News\end{tabular}}     & VT        & 0.2445 & 0.0523 & 0.1301 & 0.0603 & 0.1480 & 2.0054  & 0.1380/0.3212 & 0.4622 & -5.7674 & -7.4220  \\ 
                                                                           & VT w/o S  & 0.2555 & 0.0550 & 0.1347 & 0.0651 & 0.1491 & 2.0193  & 0.1384/0.3214 & 0.4605 & -5.2098 & -8.0488  \\ 
                                                                           & VTC       & 0.4233 & 0.1471 & 0.2059 & 0.1625 & 0.2860 & 11.3861 & 0.3778/0.4871 & 0.5955 & -6.0966 & 3.9027  \\ 
                                                                           & VTC w/o S & 0.4363 & 0.1555 & 0.2053 & 0.1698 & 0.2885 & 13.015 & 0.3967/0.5017 & 0.5916 & -6.2869 & 3.8355  \\  
                                                                           & HT        & 0.2349 & 0.0371 & 0.1352 & 0.0598 & 0.1154 & 3.5434  & 0.1097/0.3074 & 0.4987 & -5.0249 & -17.1520 \\ 
                                                                           & HT w/o S  & 0.2304 & 0.0384 & 0.1430 & 0.0580 & 0.1193 & 3.0499  & 0.1023/0.3031 & 0.4966 & -4.9433 & -16.8205 \\ \hline
\end{tabular}}
\vspace{-2mm}
\caption{Evaluation results on Multi-XScience and Multi-News datasets, both with and without the document  boundary separators. ``S" indicates document separators.} \label{tab:specialtoken}
\end{table*}

\subsection{Impact of Document Separators}

We investigate the VT, VTC, and HT models on both datasets and report the eleven evaluation metrics to explore the impact of the document boundary separators. From Table \ref{tab:specialtoken}, interestingly, we find that adding separators reduces models' performance in half of the cases (3 out of 6). For example, model VT with separators performs relatively worse on Multi-News (the results of 8 evaluation metrics are worse among 11 evaluation metrics); model VTC performs relatively worse on both Multi-XScience (the results of 9 evaluation metrics are worse among 11 evaluation metrics) and Multi-News (the results of 8 evaluation metrics are worse among 11 evaluation metrics) when with separators.
These results indicate input documents with separators are not very helpful for flat Transformer models. However, we can perceive that the HT model achieves better performance on both datasets with document boundary separators.

Another interesting finding is the most commonly used ROUGE, in a few cases, shows the opposite result from other evaluation metrics.
For instance, on the Multi-XScience dataset, the VT (with document boundary separators) shows better ROUGE results than VT (without document boundary separators) but contradicts the results on “R-WE", “BLEU”, “S\textsuperscript{3}”, ”BertScore", “Redundancy” and “Relevance". 
It indicates that the ROUGE-centric evaluation system needs to be updated and the measurement of summarization can not rely solely on ROUGE.

\begin{figure}[h]
    \centering
    \includegraphics[width=0.39\textwidth]{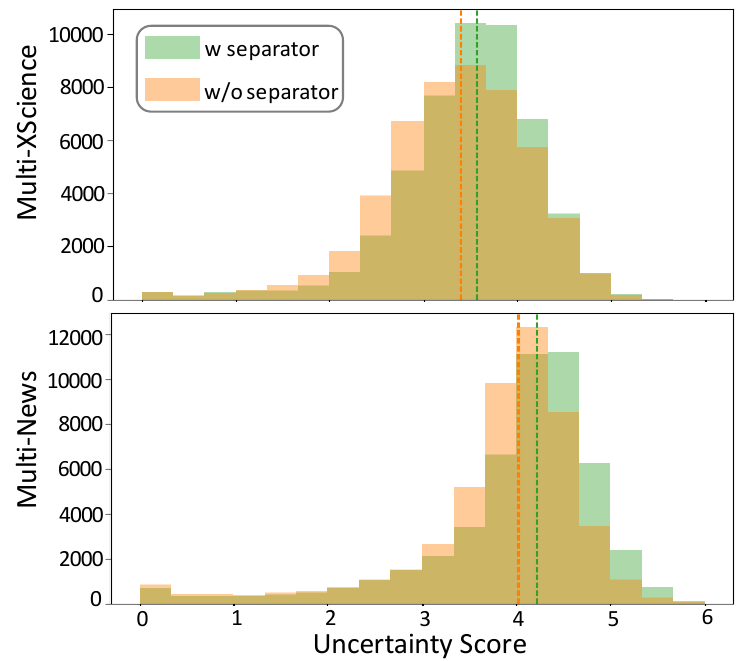}
    \vspace{-2mm}
    \caption{The uncertainty scores of VTC on Multi-News and Multi-XScience. The x-axis and y-axis are the value of uncertainty scores and the number of tokens.}
    \label{fig:all_entropy}
\end{figure}

We also discover the relations between document boundary separators and token uncertainty scores. Figure \ref{fig:all_entropy} shows the uncertainty scores of generated tokens of VTC models on both datasets. Surprisingly, the figure reflects that separators are associated with high uncertainty score actions which means the separators increase the predictive uncertainty of models. Possible because the separators have no semantic relations with the source documents and separators may be regarded as noise to increase the predictive uncertainty. The median uncertainty score of the Multi-News is larger than the Multi-XScience aligning with the size of datasets.

\begin{figure}[t]
  \centering
  \subfigure{
\begin{minipage}[t]{0.45\linewidth}
\centering
\includegraphics[width=1\textwidth]{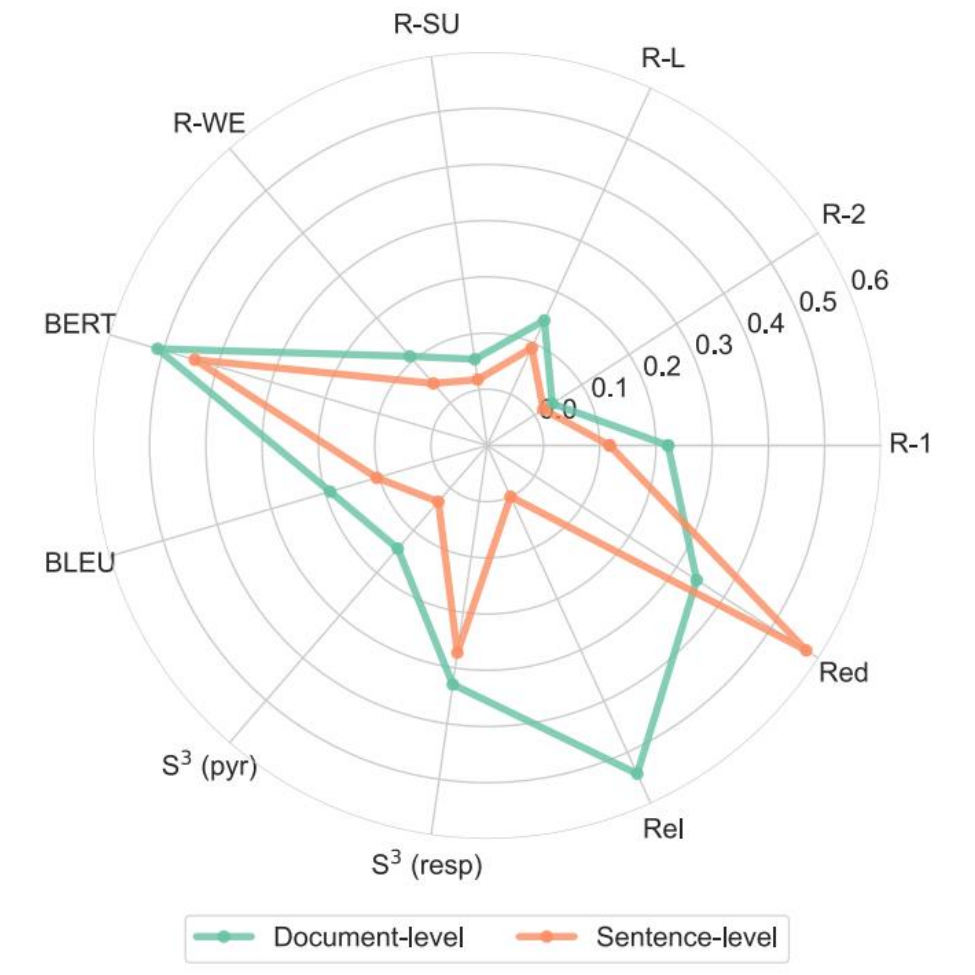}
\label{fig:MultiXScience_without_sepetators}
\end{minipage}}
\subfigure{
\begin{minipage}[t]{0.46\linewidth}
\centering
\includegraphics[width=1\textwidth]{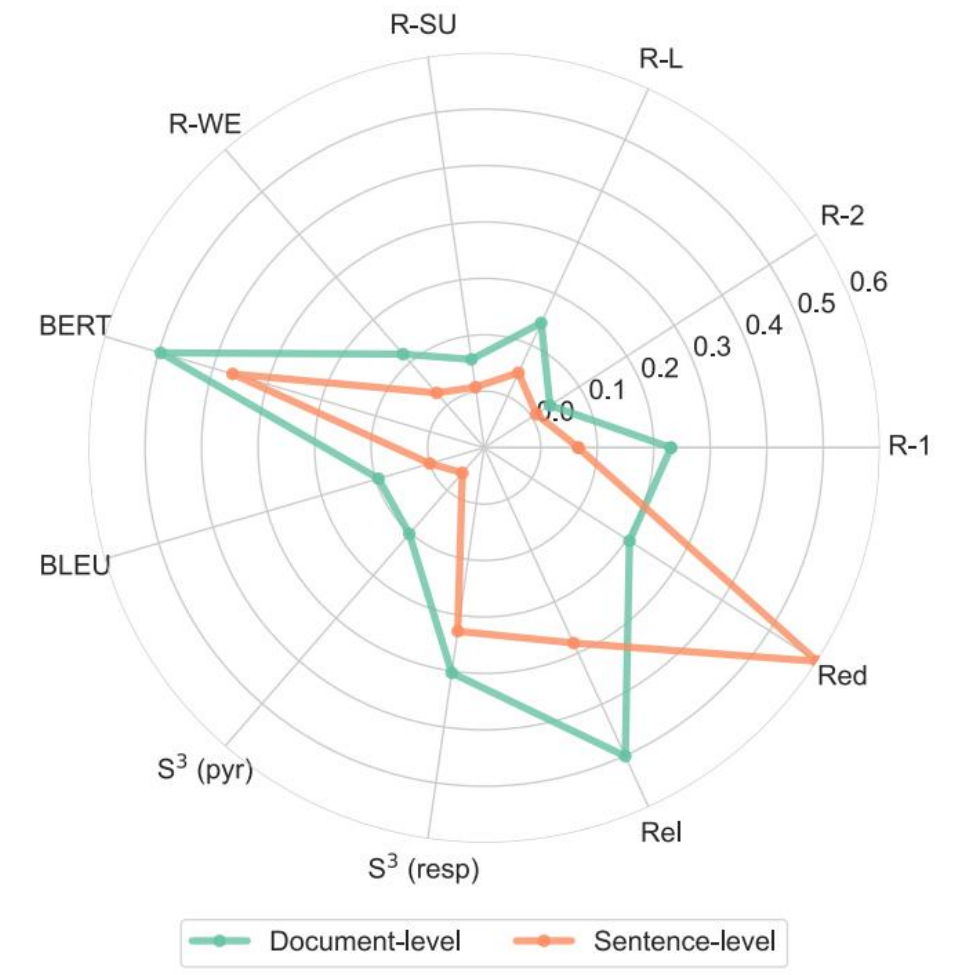}
\label{fig:Multi-News_without_sepetators}
\end{minipage}}
\vspace{-4mm}
  \caption{Performance variation with document-level (green line) and sentence-level (orange line) HT models on Multi-XScience (left) and Multi-News (right) datasets. BLEU, Redundancy and Relevance are scaled (0 to 0.6) to make all point in the plot boundary.}
\label{fig:doc_sent}
\end{figure}

\subsection{Quantitative Performance on Different Transformer Structures}

We investigate (1) the effectiveness of different Transformer architectures: flat Transformer (VT,VTC) and hierarchical Transformer (HT); (2) the influences of different granularities within hierarchical Transformer structure. The results are also found in Table \ref{tab:specialtoken}. In most evaluation metrics, the HT model can not achieve as good results as two flat Transformer models on both datasets. The two potential reasons are: (1) the pipeline of the HT model is longer than the flat Transformer models which makes the HT model hard to train. (2) the Multi-XScience and Multi-New datasets are not long document summarization datasets. The average document length of Multi-XScience and Multi-New are 778.08 and 2103.49. From the experimental results, we can conclude that the HT model is more suitable for lengthy documents, implying that flat Transformer models are a good choice for tasks with shorter documents.

As mentioned in Section \ref{sec: transformer structures}, to evaluate the influences of different granularities within the hierarchical Transformer structure, we modify the local Transformer layers to encode individual documents. Figure \ref{fig:doc_sent} shows the performances of document-level and sentence-level HT models.
All the metrics are showing better performances with the document-level HT compared to the sentence-level HT as the green line exceeds the boundary of the orange line in every dimension (redundancy is the lower the better). The apparent trend implies that a higher level of granularity is more favorable for the hierarchical Transformer structure.

\subsection{Quantitative Performance on the Sensitivity of Encoder and Decoder}

\begin{table*} \small
\centering
\resizebox{0.9\textwidth}{!}{%
\begin{tabular}{|c|c||c|c|c|c|c|c|c|c|c|c|}
\hline
Datasets                       & Models                       & R-1$\uparrow$ & R-2$\uparrow$ & R-L$\uparrow$ & R-SU$\uparrow$& R-WE$\uparrow$ & BLEU$\uparrow$ & S\textsuperscript{3} (pyr/resp)$\uparrow$ & BS$\uparrow$ & Red$\downarrow$ & Rel$\uparrow$ \\ \hline \hline
\multirow{6}{*}{\begin{tabular}[c]{@{}c@{}}Multi\\ -XScience\end{tabular}} & En ($\alpha$=$1\text{e-}3$)  & 0.2656  & 0.0477  & 0.1507  & 0.0739   & 0.1660   & 4.6288  & 0.2560/0.3881   & 0.5593    & -5.2615    & 2.5252    \\ 
                                                                           & De ($\alpha$=$1\text{e-}3$)  & 0.2637  & 0.0483  & 0.1499  & 0.0735   & 0.1676   & 4.8116  & 0.2573/0.3890   & 0.5608    & -5.2806    & 2.5377    \\   
                                                                           & En ($\alpha$=$1\text{e-}2$)  & 0.2433  & 0.0412  & 0.1386  & 0.0650   & 0.1523   & 4.0228  & 0.2276/0.3713   & 0.5506    & -5.2222    & 2.4878    \\   
                                                                           & De ($\alpha$=$1\text{e-}2$)  & 0.2130  & 0.0362  & 0.1277  & 0.0512   & 0.1333   & 2.6732  & 0.1933/0.3535   & 0.5189    & -4.5961    & 2.4406    \\   
                                                                           & En ($\alpha$=$1\text{e-}1$)  & 0.0305  & 0.0019  & 0.0232  & 0.0035   & 0.0057   & 0.0979  & -0.0786/0.2085  & 0.3631    & -1.8267    & 0.1420    \\   
                                                                           & De ($\alpha$=$1\text{e-}1$)  & 0.0282  & 0.0039  & 0.0259  & 0.0019   & 0.0012   & 0.0422  & -0.0350/0.2347  & 0.3533    & -0.9935    & 1.2109    \\ \hline \hline
\multirow{6}{*}{\begin{tabular}[c]{@{}c@{}}Multi\\ -News\end{tabular}}     & En ($\alpha$=$1\text{e-}3$)  & 0.4178  & 0.1439  & 0.2063  & 0.1598   & 0.2817   & 10.5326 & 0.3345/0.4623   & 0.5943    & -5.8867    & 3.8567    \\  
                                                                           & De  ($\alpha$=$1\text{e-}3$) & 0.4172  & 0.1427  & 0.2053  & 0.1589   & 0.2802   & 10.6737 & 0.3348/0.4625   & 0.5941    & -5.8923    & 3.8533    \\   
                                                                           & En ($\alpha$=$1\text{e-}2$)  & 0.2899 & 0.0689  & 0.1405  & 0.0888   & 0.2095   & 5.5596  & 0.2260/0.3778   & 0.5335    & -5.2695    & 3.7854    \\   
                                                                           & De ($\alpha$=$1\text{e-}2$)  & 0.2248  & 0.0602  & 0.1134  & 0.0706   & 0.1842   & 4.0850  & 0.2288/0.3793   & 0.4972    & -4.7247    & 3.3888    \\   
                                                                           & En ($\alpha$=$1\text{e-}1$)  & 0.0938  & 0.0049  & 0.0724  & 0.0101   & 0.0266   & 0.0549  & -0.0499/ 0.2223 & 0.3330    & -1.2151    & 1.7586    \\   
                                                                           & De ($\alpha$=$1\text{e-}1$)  & 0.0458  & 0.0018  & 0.0330  & 0.0041   & 0.0186   & 0.0476  & -0.0537/0.2207  & 0.3410    & -2.3011    & 1.3539    \\ \hline
\end{tabular}
}
\vspace{-2mm}
\caption{Evaluation results on Multi-XScience and Multi-News datasets about the encoder-decoder structure.}
\label{encoder_decoder}
\end{table*}

To investigate the hypothesis in section \ref{sec: encoder and decoder}, we select the VTC model as the foundation for evaluating the effectiveness of the encoder-decoder structure on the Multi-XScience and Multi-News datasets. 
By examining Table \ref{encoder_decoder}, we observe large differences in performance when introducing noise to the encoder and decoder in highly noisy scenarios (with $\alpha=1\text{e-}1$ and $\alpha=1\text{e-}2$). Specifically, in noisy conditions, we find that adding noise to the decoder has a more substantial impact on performance compared to adding noise to the encoder. However, as the noise levels decreased, the performance gaps between the two approaches narrowed. This observation supports our initial hypothesis that the decoder is more sensitive than the encoder. The potential reasons are: (1) errors or inaccuracies in the decoder can have a cascading effect on subsequent tokens generated during decoding. This error propagation phenomenon can make the decoder more sensitive to small perturbations, as any mistakes or noise introduced during decoding can amplify and affect the overall quality of the generated summary; (2) Transformer-based models often employ an attention mechanism that allows the decoder to focus on different parts of the encoded input during the decoding process. The decoder's sensitivity is crucial in effectively attending to relevant information, and even slight perturbations in the encoded input can impact the attention weights and subsequently influence the decoding process. Consequently, it underscores the crucial role played by the decoder in summarization tasks. These findings shed light on the high importance of the decoder's contribution to the overall summarization process.

\subsection{Quantitative Performance of Different Training Strategies}

\begin{figure}[h]
  \centering
  \subfigure{
\begin{minipage}[t]{0.48\linewidth}
\centering
\includegraphics[width=1\textwidth]{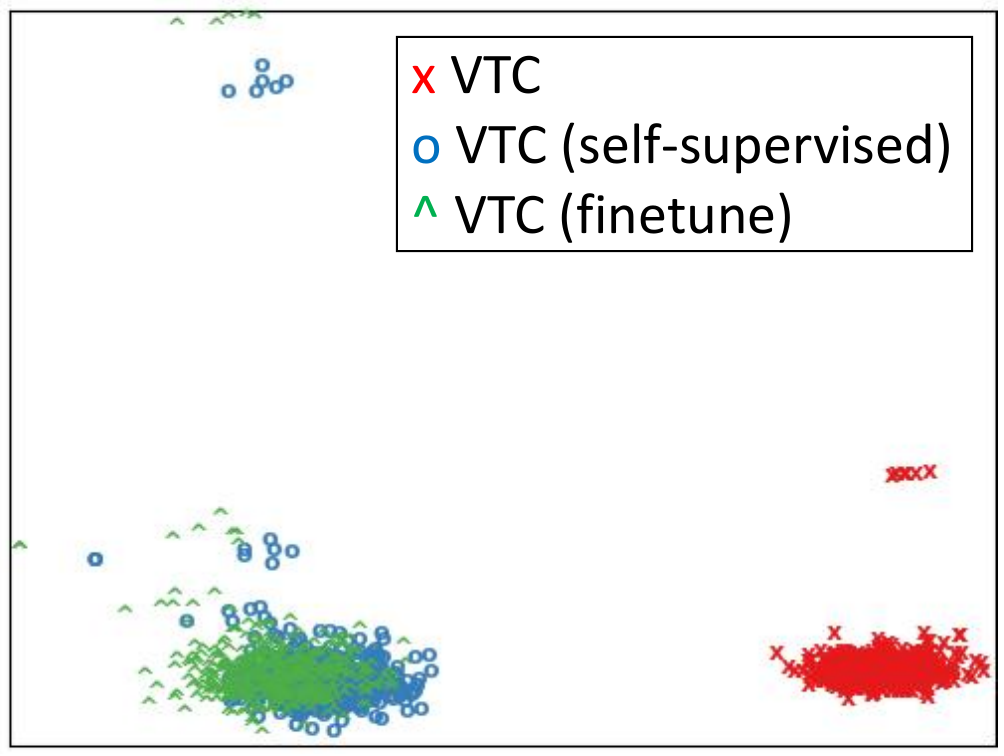}
Multi-News
\label{fig:PCA_Multi-News}
\end{minipage}}
\subfigure{
\begin{minipage}[t]{0.48\linewidth}
\centering
\includegraphics[width=1\textwidth]{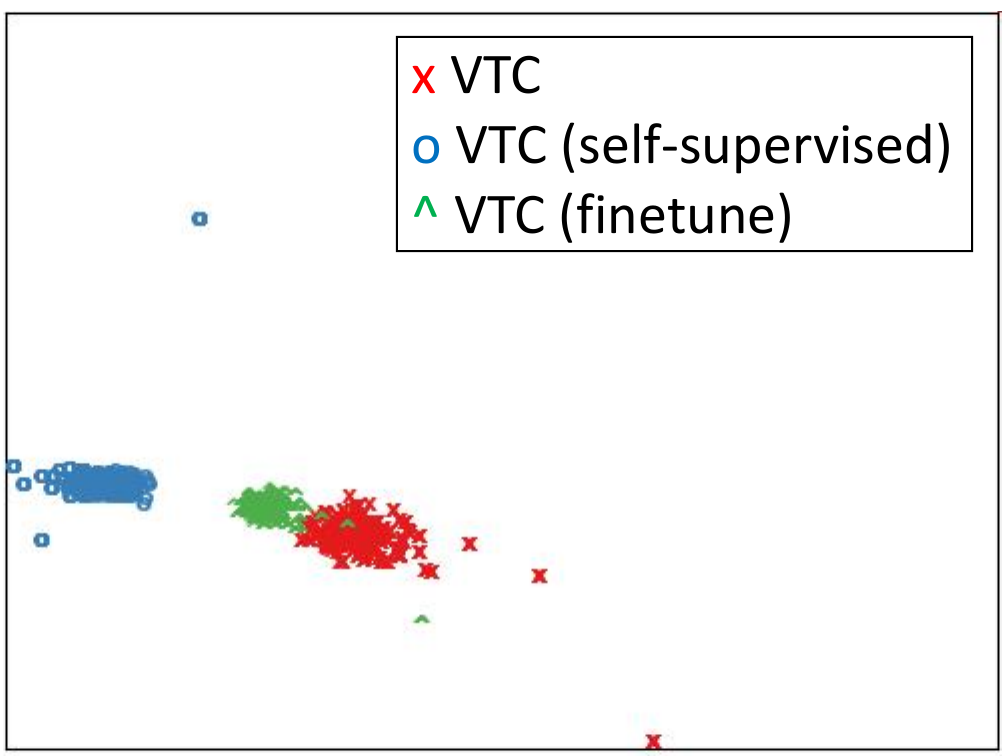}
Multi-XScience
\label{fig:PCA_MultiXScience}
\end{minipage}}
\vspace{-2mm}
\caption{The feature visualization of VTC, VTC with self-supervised training and VTC with finetuning after self-supervised training with PCA.}
\label{fig:pca-vis}
\end{figure}

The experimental results presented in Table \ref{tab: pretraining} provide an overview of the performance of the VTC model trained using different pretraining strategies on the Multi-XScience and Multi-News datasets. In the table, the VTC is trained on the original document set and golden summary pairs. The ``finetune" strategy refers to the training of the model on the pseudo dataset (introduced in Section \ref{sec: training strategies}) first and then fine-tuning on the original dataset. The ``self-supervised" strategy denotes training the VTC model exclusively on the pseudo dataset. The ``mix" strategy illustrates training the model using a combination of the pseudo dataset and the original dataset. By comparing the results obtained from these different training strategies, we aim to identify the most effective approach for each dataset.

For the Multi-XScience, the results show that the VTC (pretrain-finetune) strategy outperforms the VTC trained on the original dataset across most metrics, indicating the effectiveness of the pretrain-finetune strategy in improving summarization quality. In contrast, the VTC (self-supervised) exhibits lower performance compared to the VTC (pretrain-finetune), suggesting that just self-supervised training is less effective for this dataset.

Similarly, for the Multi-News dataset, the results imply the VTC model achieves good performance across all metrics, with higher scores on the VTC (pretrain-finetune) strategy, showcasing improved summarization quality. Conversely, the VTC (self-supervised) and VTC (mix)  strategy yields lower performance compared to the other strategies.

\begin{table*} \small
\centering
\resizebox{0.9\textwidth}{!}{%
\begin{tabular}{|c|c||c|c|c|c|c|c|c|c|c|c|}
\hline
Datasets                       & Models                       & R-1$\uparrow$ & R-2$\uparrow$ & R-L$\uparrow$ & R-SU$\uparrow$& R-WE$\uparrow$ & BLEU$\uparrow$ & S\textsuperscript{3} (pyr/resp)$\uparrow$ & BS$\uparrow$ & Red$\downarrow$ & Rel$\uparrow$ \\ \hline \hline
\multirow{4}{*}{\begin{tabular}[c]{@{}l@{}}Multi-\\ XScience\end{tabular}} &
  VTC &
  0.2635 &
  0.0483 &
  0.1499 &
  0.0734 &
  \multicolumn{1}{r|}{0.1659} &
  \multicolumn{1}{r|}{4.6037} &
  0.2561/0.3885 &
  \multicolumn{1}{r|}{0.5590} &
  \multicolumn{1}{r|}{-7.0585} &
  \multicolumn{1}{r|}{-4.5802} \\   
 &
  VTC (finetune) &
  0.2955 &
  0.0558 &
  0.1671 &
  0.0879 &
  0.1770 &
  3.9727 &
  0.2569/0.3886 &
  0.5511 &
  -5.0020 &
  2.5824
   \\   
 &
  VTC(self-supervised) &
  0.2585 &
  0.0368 &
  0.1471 &
  0.0678 &
  0.1325 &
  1.2885 &
  0.1694/0.3343 &
  0.5173 &
  -5.3546 &
  2.2064
   \\   
 &
  VTC(mix) &
  0.2547 &
  0.0350 &
  0.1468 &
  0.0653 &
  0.1324 &
  1.2922 &
  0.1526/0.3246 &
  0.5176 &
  -5.3285 &
  2.1945
   \\ \hline \hline
\multirow{4}{*}{\begin{tabular}[c]{@{}l@{}}Multi-\\ News\end{tabular}} &
  VTC &
  0.4233 &
  0.1471 &
  0.2059 &
  0.1625 &
  \multicolumn{1}{r|}{0.2860} &
  \multicolumn{1}{r|}{11.3861} &
  \multicolumn{1}{r|}{0.3778/0.4871} &
  \multicolumn{1}{r|}{0.5955} &
  \multicolumn{1}{r|}{-6.0966} &
  \multicolumn{1}{r|}{3.9027} \\   
 &
  VTC (finetune) &
  \multicolumn{1}{l|}{0.4271} &
  \multicolumn{1}{l|}{0.1509} &
  \multicolumn{1}{l|}{0.2084} &
  \multicolumn{1}{l|}{0.1643} &
   0.2886 &
   11.5514 &
   0.3893/0.4960 &
   0.6004 &
   -6.2075 &
   3.9135 
   \\   
 &
  VTC(self-supervised) &
  0.2724 &
  0.0484 &
  0.1349 &
  0.0738 &
  \multicolumn{1}{r|}{0.1399} &
  \multicolumn{1}{r|}{2.8583} &
  0.1281/0.3159 &
  \multicolumn{1}{r|}{0.4737} &
  \multicolumn{1}{r|}{-5.5046} &
  \multicolumn{1}{r|}{2.4027} \\   
 &
  VTC(mix) &
  0.3046 &
  0.0673 &
  0.1485 &
  0.0938 &
  \multicolumn{1}{r|}{0.1728} &
  \multicolumn{1}{r|}{5.2611} &
  0.1909/0.3595 &
  \multicolumn{1}{r|}{0.4979} &
  \multicolumn{1}{r|}{-5.8684} &
  \multicolumn{1}{r|}{2.7281} \\ \hline
\end{tabular}}
\vspace{-2mm}
\caption{Different training strategies on Multi-News and Multi-XScience datasets.} 
\label{tab: pretraining}
\end{table*}

The comparison of these different training strategies reveals that the pretrain-finetune approach consistently leads to better summarization performance compared to the baseline VTC model and other training strategy, highlighting its effectiveness in improving summarization quality.

To find the potential reason why the finetune strategy works well, we visualize the feature distributions of three training strategies: VTC, VTC (self-supervised) VTC (finetune) using Principal Component Analysis (PCA) as illustrated in Figure \ref{fig:pca-vis}. For the Multi-News, the features come from the encoder of the VTC (self-supervised) and the VTC (finetuning) exhibits overlapping, while maintaining distance from the plain VTC. In contrast, for the Multi-XScience, the VTC (finetune) is more similar to the plain VTC but still noticeably distinct from the VTC (self-supervised). This observation is consistent with the performance results presented in Table \ref{tab: pretraining}. In the case of the Multi-XScience, finetuning the model after self-supervised training significantly improves the model's performance compared to the VTC. However, when the model is only pretrained using self-supervised learning, it performs worse than the VTC. This discrepancy can be attributed to the fact that the features of the finetuned model closely align with the VTC model's distribution since both models possess better representations for the final prediction. Conversely, for the Multi-News, the finetuned model exhibits only marginal improvements over the VTC. This observation also explains the overlap between features from the finetuned model and the self-supervised model, as finetuning adjusts the feature distribution towards the `genuine' distribution, albeit to a limited extent.

\begin{figure}[htbp]
    \centering
    \includegraphics[width=0.48\textwidth]{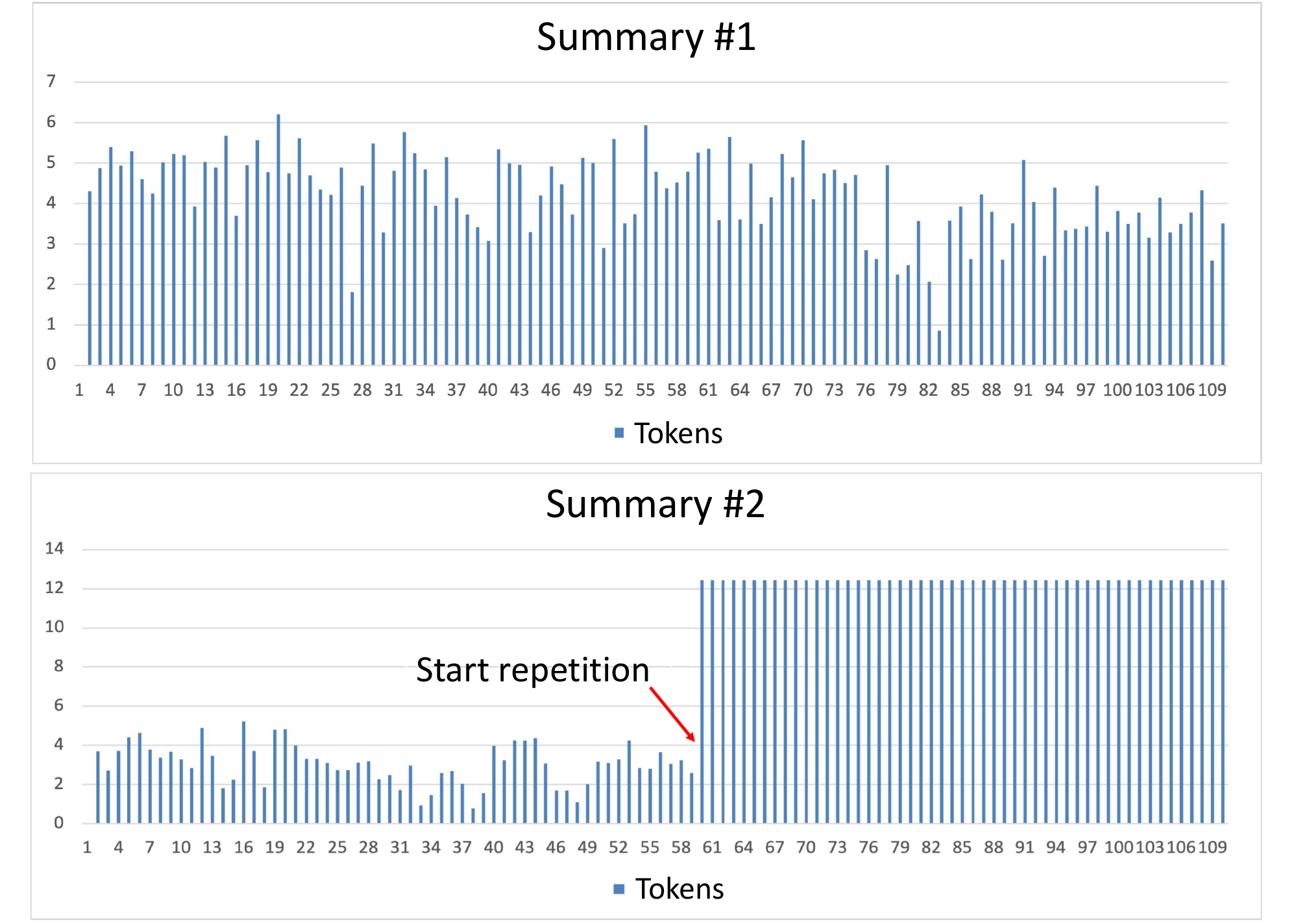}
     \caption{The relationship between uncertainty scores and token repetitions on different summaries. }
    \label{fig:repetition}
\end{figure}

\subsection{The Relation Between Repetition and Uncertainty}

We examine the correlation between repetition and uncertainty in the process of generating summaries. To assess uncertainty, we compute a score for each token generated. Two summaries are presented: one featuring repetition and the other as a standard summary without repetition. The outcomes are depicted in Figure \ref{fig:repetition}. The X-axis represents token indexes, while the Y-axis illustrates uncertainty scores for each token. In summary \#1, where no repetitions occur, the uncertainties of tokens remain within a ``normal" range. This suggests that the model successfully avoids repetitive patterns, resulting in lower uncertainty scores throughout the summary generation process. Conversely, in summaries \#2, we observe a distinct pattern. As the repetition of tokens or phrases begins, the uncertainty scores escalate rapidly. 

By comparing uncertainty scores across different time slots, we gain insights into the relationship between repetition and uncertainty in abstractive summarization. When a repetition phenomenon occurs, we observe notable changes in the uncertainty score, indicating a correlation between the two factors. Specifically, as the model generates repetitive sentences or words, the uncertainty score tends to increase. This increase in uncertainty suggests that the model becomes less confident and more uncertain about the appropriateness or relevance of the repeated elements within the summary.
By understanding this relationship, we can devise strategies to mitigate repetition and subsequently enhance the quality of generated summaries. By reducing uncertainty through the minimization of repetition, we pave the way for more accurate and reliable abstractive summarization.

\section{Conclusion and Discussion}

This study attempts to empirically examine the influences on Transformer behaviors from five important perspectives: document boundary separators, Transformer structures, the sensitivity of encoder-decoder architecture, training strategies, and the relationship between repetition and uncertainty in generated
summaries. 
We first explore the impact of separators on two flat Transformer and one hierarchical Transformer structure.

Experiments indicate that adding separators makes hierarchical Transformers aware of document boundaries, unlike flat Transformers. This suggests that for models handling complex structures, separators can enhance performance. The necessity of adopting separators should be considered depending on the Transformer structure applied.

The Transformer structure exploring experiments demonstrate that a higher level of granularity is favorable for the hierarchical Transformer structure. The experiments also demonstrate the simple structure, flat Transformer, has been able to show better performance on the Multi-XScience and Multi-News datasets than the complicated hierarchical Transformer structure. The flat Transformer models are sufficient for MDS tasks with relatively short length of documents. 


Furthermore, adding noise to the decoder affects performance more than adding noise to the encoder. This sensitivity is likely due to error propagation during decoding and the attention mechanism's dependence on accurate encoding. These results emphasize the decoder's crucial role in producing high-quality summaries and its significant impact on the summarization process.

The pretrain-finetune strategy that trains the model on the pseudo labels first and then fine-tuning it on the original dataset consistently leads to improved summarization performance when compared to other training strategies. This finding highlights the effectiveness of the pretrain-finetune strategy in enhancing MDS model performance.

Moreover, the analysis of the relations between repetition and uncertainty provides valuable insights into improving the quality of generated summaries. The findings suggest that as repetition occurs in the summaries, there is a noticeable increase in uncertainty scores. By recognizing this relationship, strategies can be developed to mitigate repetition and reduce uncertainty, ultimately enhancing the overall quality of abstractive summaries. These insights contribute to the advancement of abstractive summarization techniques and open avenues for further research in improving the reliability and effectiveness of summary generation.

We also point out the possible exploring direction for future MDS work: (1) evaluate the generated summaries from multiple evaluations; (2) add the higher level of granularity information into the models; (3) investigate the MDS method for particularly long input documents; (4) pay more attention to the decoder when designing the Transformer-based summarization models; (5) try to reduce the Sudden sharp increase and high uncertainty score during the summary generation process.

\section*{Limitations}
The original Hierarchical Transformer (HT) model is trained on four GPUs (NVIDIA TITAN Xp) for 500, 000 steps, but with an unspecified batch-size. In order to keep a fair comparison and consider the limitation of our computation resource, all the models reported in the paper are trained on the same one GPU, which in turn influences the setting of batch-size. It may effect the performance of HT model.

\bibliography{custom}

\begin{thebibliography}{26}
\expandafter\ifx\csname natexlab\endcsname\relax\def\natexlab#1{#1}\fi

\bibitem[{Brazinskas et~al.(2020)Brazinskas, Lapata, and Titov}]{Brazinskas2020few}
Arthur Brazinskas, Mirella Lapata, and Ivan Titov. 2020.
\newblock \href {https://arxiv.org/pdf/2004.14884v2.pdf} {Few-shot learning for opinion summarization}.
\newblock In \emph{Proceedings of the 2020 Conference on Empirical Methods in Natural Language Processing (EMNLP 2020)}, pages 4119--4135, Online.

\bibitem[{Chu and Liu(2019)}]{chu2019meansum}
Eric Chu and Peter~J. Liu. 2019.
\newblock \href {https://arxiv.org/pdf/1810.05739.pdf} {Meansum: A neural model for unsupervised multi-document abstractive summarization}.
\newblock In \emph{Proceedings of the 36th International Conference on Machine Learning (ICML 2019)}, pages 1223--1232, Long Beach, {United States}.

\bibitem[{Fabbri et~al.(2019)Fabbri, Li, She, Li, and Radev}]{Alexander2019multinews}
Alexander~R. Fabbri, Irene Li, Tianwei She, Suyi Li, and Dragomir~R. Radev. 2019.
\newblock \href {https://doi.org/10.18653/v1/p19-1102} {{Multi-News: A Large-Scale Multi-Document Summarization Dataset and Abstractive Hierarchical Model}}.
\newblock In \emph{Proceedings of the 57th Conference of the Association for Computational Linguistics (ACL 2019)}, pages 1074--1084, Florence, Italy.

\bibitem[{Jin and Wan(2020)}]{han2020abstractive}
Hanqi Jin and Xiaojun Wan. 2020.
\newblock \href {https://doi.org/10.18653/v1/2020.findings-emnlp.231} {{Abstractive Multi-Document Summarization via Joint Learning with Single-Document Summarization}}.
\newblock In \emph{Findings of the Association for Computational Linguistics (EMNLP 2020)}, pages 2545--2554, online.

\bibitem[{Li et~al.(2020)Li, Xiao, Liu, Wu, Wang, and Du}]{li2020leveraging}
Wei Li, Xinyan Xiao, Jiachen Liu, Hua Wu, Haifeng Wang, and Junping Du. 2020.
\newblock \href {https://doi.org/10.18653/v1/2020.acl-main.555} {{Leveraging Graph to Improve Abstractive Multi-Document Summarization}}.
\newblock In \emph{Proceedings of the 58th Annual Meeting of the Association for Computational Linguistics (ACL 2020)}, pages 6232--6243, Online.

\bibitem[{Lin(2004)}]{lin-2004-rouge}
Chin-Yew Lin. 2004.
\newblock \href {https://aclanthology.org/W04-1013} {{ROUGE: A Package for Automatic Evaluation of Summaries}}.
\newblock In \emph{Proceedings of the Workshop of Text Summarization Branches Out}, pages 74--81, Barcelona, Spain.

\bibitem[{Lin et~al.(2021)Lin, Wang, Liu, and Qiu}]{lin2021survey}
Tianyang Lin, Yuxin Wang, Xiangyang Liu, and Xipeng Qiu. 2021.
\newblock \href {https://arxiv.org/abs/2106.04554} {{A Survey of Transformers}}.
\newblock volume abs/2106.04554.

\bibitem[{Liu et~al.(2018)Liu, Saleh, Pot, Goodrich, Sepassi, Kaiser, and Shazeer}]{liu2018generating}
Peter~J. Liu, Mohammad Saleh, Etienne Pot, Ben Goodrich, Ryan Sepassi, Lukasz Kaiser, and Noam Shazeer. 2018.
\newblock \href {https://openreview.net/forum?id=Hyg0vbWC-} {{Generating Wikipedia by Summarizing Long Sequences}}.
\newblock In \emph{Proceedings of the 6th International Conference on Learning Representations (ICLR 2018)}, Vancouver, Canada.

\bibitem[{Liu et~al.(2021)Liu, Cao, Yang, and Wen}]{liu2021highlight}
Shuaiqi Liu, Jiannong Cao, Ruosong Yang, and Zhiyuan Wen. 2021.
\newblock \href {https://doi.org/10.18653/v1/2021.findings-acl.445} {{Highlight-Transformer: Leveraging Key Phrase Aware Attention to Improve Abstractive Multi-Document Summarization}}.
\newblock In \emph{Findings of the Association for Computational Linguistics (ACL/IJCNLP 2021)}, pages 5021--5027, Online.

\bibitem[{Liu and Lapata(2019)}]{liu-lapata-2019-hierarchical}
Yang Liu and Mirella Lapata. 2019.
\newblock \href {https://aclanthology.org/P19-1500} {{Hierarchical Transformers for Multi-Document Summarization}}.
\newblock In \emph{Proceedings of the 57th Conference of the Association for Computational Linguistics (ACL 2019)}, pages 5070--5081, Florence, Italy.

\bibitem[{Liu et~al.(2022)Liu, Liu, Radev, and Neubig}]{liu2022BRIO}
Yixin Liu, Pengfei Liu, Dragomir~R. Radev, and Graham Neubig. 2022.
\newblock \href {https://aclanthology.org/2022.acl-long.207} {{BRIO: Bringing Order to Abstractive Summarization}}.
\newblock In \emph{Proceedings of the 60th Annual Meeting of the Association for Computational Linguistics (ACL 2022)}, pages 2890--2903, Dublin, Ireland.

\bibitem[{Liu et~al.(2023)Liu, Chen, Luo, and Zhu}]{liu2023reducing}
Yizhu Liu, Xinyue Chen, Xusheng Luo, and Kenny~Q. Zhu. 2023.
\newblock \href {https://www.cambridge.org/core/journals/natural-language-engineering/article/abs/reducing-repetition-in-convolutional-abstractive-summarization/2247BFF85CEFD40C1189B05CD6E9BE85} {Reducing repetition in convolutional abstractive summarization}.
\newblock volume~29, pages 81--109.

\bibitem[{Lu et~al.(2020)Lu, Dong, and Charlin}]{Yao2020multi}
Yao Lu, Yue Dong, and Laurent Charlin. 2020.
\newblock \href {https://doi.org/10.18653/v1/2020.emnlp-main.648} {{Multi-XScience: A Large-scale Dataset for Extreme Multi-document Summarization of Scientific Articles}}.
\newblock In \emph{Proceedings of the 2020 Conference on Empirical Methods in Natural Language Processing (EMNLP 2020)}, pages 8068--8074, Online.

\bibitem[{Ma et~al.(2022{\natexlab{a}})Ma, Zhang, Guo, Wang, and Sheng}]{ma2020multi}
Congbo Ma, Wei~Emma Zhang, Mingyu Guo, Hu~Wang, and QUAN~Z. Sheng. 2022{\natexlab{a}}.
\newblock \href {https://doi.org/10.1145/3529754} {{Multi-Document Summarization via Deep Learning Techniques: A Survey}}.
\newblock \emph{ACM Computing Surveys}.

\bibitem[{Ma et~al.(2022{\natexlab{b}})Ma, Zhang, Pitawela, Qu, Zhuang, and Wang}]{ma2022document}
Congbo Ma, Wei~Emma Zhang, Pitawelayalage Dasun~Dileepa Pitawela, Yutong Qu, Haojie Zhuang, and Hu~Wang. 2022{\natexlab{b}}.
\newblock \href {https://arxiv.org/pdf/2209.05929.pdf} {Document-aware positional encoding and linguistic-guided encoding for abstractive multi-document summarization}.
\newblock In \emph{IEEE World Congress On Computational Intelligence (WCCI 2022)}, Padua, Italy.

\bibitem[{Mao et~al.(2020)Mao, Qu, Xie, Ren, and Han}]{mao2020multi}
Yuning Mao, Yanru Qu, Yiqing Xie, Xiang Ren, and Jiawei Han. 2020.
\newblock \href {https://doi.org/10.18653/v1/2020.emnlp-main.136} {Multi-document summarization with maximal marginal relevance-guided reinforcement learning}.
\newblock In \emph{Proceedings of the 2020 Conference on Empirical Methods in Natural Language Processing (EMNLP 2020)}, pages 1737--1751, Online.

\bibitem[{Ng and Abrecht(2015)}]{ng-abrecht-2015-better}
Jun{-}Ping Ng and Viktoria Abrecht. 2015.
\newblock \href {https://doi.org/10.18653/v1/d15-1222} {{Better Summarization Evaluation with Word Embeddings for ROUGE}}.
\newblock In \emph{Proceedings of the 2015 Conference on Empirical Methods in Natural Language Processing (EMNLP 2015)}, pages 1925--1930, Lisbon, Portugal.

\bibitem[{Papineni et~al.(2002)Papineni, Roukos, Ward, and Zhu}]{Papineni-etal}
Kishore Papineni, Salim Roukos, Todd Ward, and Wei-Jing Zhu. 2002.
\newblock \href {https://doi.org/10.3115/1073083.1073135} {{BLEU: A Method for Automatic Evaluation of Machine Translation}}.
\newblock In \emph{Proceedings of the 40th Annual Meeting of the Association for Computational Linguistics (ACL 2002)}, pages 311--318, Philadelphia, United States.

\bibitem[{Pasunuru et~al.(2021)Pasunuru, Liu, Bansal, Ravi, and Dreyer}]{pasunuru2021efficiently}
Ramakanth Pasunuru, Mengwen Liu, Mohit Bansal, Sujith Ravi, and Markus Dreyer. 2021.
\newblock \href {https://doi.org/10.18653/v1/2021.naacl-main.380} {{Efficiently Summarizing Text and Graph Encodings of Multi-Document Clusters}}.
\newblock In \emph{Proceedings of the 2021 Conference of the North American Chapter of the Association for Computational Linguistics: Human Language Technologies (NAACL-HLT 2021)}, pages 4768--4779, Online.

\bibitem[{Peyrard(2019)}]{Simple-Theoretical}
Maxime Peyrard. 2019.
\newblock \href {https://doi.org/10.18653/v1/p19-1101} {{A Simple Theoretical Model of Importance for Summarization}}.
\newblock In \emph{Proceedings of the 57th Conference of the Association for Computational Linguistics (ACL 2019)}, pages 1059--1073, Florence, Italy.

\bibitem[{Peyrard et~al.(2017)Peyrard, Botschen, and Gurevych}]{peyrard-etal-2017-learning}
Maxime Peyrard, Teresa Botschen, and Iryna Gurevych. 2017.
\newblock \href {https://doi.org/10.18653/v1/w17-4510} {{Learning to Score System Summaries for Better Content Selection Evaluation}}.
\newblock In \emph{Proceedings of the Workshop on New Frontiers in Summarization (NFiS@EMNLP 2017)}, pages 74--84, Copenhagen, Denmark.

\bibitem[{Vaswani et~al.(2017)Vaswani, Shazeer, Parmar, Uszkoreit, Jones, Gomez, Kaiser, and Polosukhin}]{Attention-is-all-uneed}
Ashish Vaswani, Noam Shazeer, Niki Parmar, Jakob Uszkoreit, Llion Jones, Aidan~N Gomez, {\L}ukasz Kaiser, and Illia Polosukhin. 2017.
\newblock \href {https://proceedings.neurips.cc/paper/2017/hash/3f5ee243547dee91fbd053c1c4a845aa-Abstract.html} {{Attention Is All You Need}}.
\newblock In \emph{Proceedings of the 31st Annual Conference on Neural Information Processing Systems (NIPS 2017)}, pages 5998--6008, Long Beach, United States.

\bibitem[{Xiao et~al.(2022)Xiao, Beltagy, Carenini, and Cohan}]{wen2022PRIMERA}
Wen Xiao, Iz~Beltagy, Giuseppe Carenini, and Arman Cohan. 2022.
\newblock \href {https://aclanthology.org/2022.acl-long.360} {{PRIMERA: Pyramid-based Masked Sentence Pre-training for Multi-document Summarization}}.
\newblock In \emph{Proceedings of the 60th Annual Meeting of the Association for Computational Linguistics (ACL 2022)}, pages 5245--5263, Dublin, Ireland.

\bibitem[{Xu et~al.(2020)Xu, Desai, and Durrett}]{xu2020understanding}
Jiacheng Xu, Shrey Desai, and Greg Durrett. 2020.
\newblock \href {https://doi.org/10.18653/v1/2020.emnlp-main.508} {Understanding neural abstractive summarization models via uncertainty}.
\newblock In \emph{Proceedings of the 2020 Conference on Empirical Methods in Natural Language Processing (EMNLP 2020)}, pages 6275--6281, Online.

\bibitem[{Zhang et~al.(2020)Zhang, Kishore, Wu, Weinberger, and Artzi}]{Zhang2020BERTScore}
Tianyi Zhang, Varsha Kishore, Felix Wu, Kilian~Q. Weinberger, and Yoav Artzi. 2020.
\newblock \href {https://openreview.net/forum?id=SkeHuCVFDr} {{BERTScore: Evaluating Text Generation with BERT}}.
\newblock In \emph{Proceedings of the 8th International Conference on Learning Representations (ICLR 2020)}, Addis Ababa, Ethiopia.

\bibitem[{Zhao et~al.(2022)Zhao, Huang, Chowdhury, Chandrasekaran, McKeown, and Chaturvedi}]{Chao2022Read}
Chao Zhao, Tenghao Huang, Somnath Basu~Roy Chowdhury, Muthu~Kumar Chandrasekaran, Kathleen~R. McKeown, and Snigdha Chaturvedi. 2022.
\newblock \href {https://doi.org/10.18653/v1/2022.findings-acl.51} {Read top news first: {A} document reordering approach for multi-document news summarization}.
\newblock In \emph{Findings of the Association for Computational Linguistics (ACL 2022 findings)}, pages 613--621, Dublin, Ireland.

\end{thebibliography}
\appendix

\section{Appendix}
\label{sec:appendix}

\subsection{Implementation Details}

The training of all models begins with an initial learning rate of 2. An initial warm-up phase spans the first 8,000 steps, followed by a subsequent multi-step learning rate reduction. During the training process, a batch size of 4,096 is utilized, and the optimization is performed for 20,000 steps using the \textit{Adam} optimizer. A dropout rate of 0.2 is employed to enhance model robustness. All experiments are conducted on a single NVIDIA 3090 GPU with one Intel i9-10900X CPU. The operating environment is provided by Ubuntu 22.04.3 LTS.

\subsection{Summarization Models}
\vspace{1mm}
\noindent \textbf{Vanilla Transformer (VT)} \cite{Attention-is-all-uneed} is a sequence-to-sequence model that is proposed for machine translation task. It is subsequently generalized in various tasks of NLP due to its strong performance \cite{lin2021survey}.

\vspace{1mm}
\noindent \textbf{Vanilla Transformer with Copy Mechanism (VTC)\footnote{We implement the VT and VTC based on https://github.com/Alex-Fabbri/Multi-News/tree/master/code/OpenNMT-py-baselines.}. }
This variant has a mechanism to copy the attention distribution that one of the randomly chosen attention heads from the encoder side into the decoder so that the generated text becomes less repetitive and less factually inaccurate. 

\vspace{1mm}
\noindent \textbf{Hierarchical Transformer (HT)}
\cite{liu-lapata-2019-hierarchical} proposed a hierarchical attention structure to attend long sequences effectively and capture cross-paragraph contextual relationships. The local Transformer layers encode individual paragraphs and global Transformer layers exchange paragraph-level information from local layers across paragraphs.

\subsection{Datasets}
The empirical studies are based on two widely used MDS datasets: Multi-XScience \cite{Yao2020multi} and Multi-News \cite{Alexander2019multinews}. Multi-XScience contains data from scientific articles. The task of this dataset is to generate the related work section of a target paper based on its abstract and the abstracts of the articles it refers to. Multi-News collects news articles from the site "newser.com." Each set of source documents has a professionally written summary and the task is to generate that summary based on the sources. Table \ref{tab: dataset} describes the statistics of these two datasets, including the size of the train, test, and validation set, the average document length, and the average summary length.

\begin{table*}[]
\centering
\scalebox{0.7}{
\begin{tabular}{|c|c|c|c|}
\hline
Datasets       & Train/ Test/ Validation & \begin{tabular}[c]{@{}c@{}}Average  Document \\ Length\end{tabular} & \begin{tabular}[c]{@{}c@{}}Average  Summary \\Length\end{tabular} \\ \hline
Multi-XScience & 30,369 / 5,093/ 5,066   & 778.08                                                            & 116.44                                                           \\ \hline
Multi-News     & 44,972 / 5,622 / 5,622  & 2,103.49                                                          & 263.66                                                           \\ \hline
\end{tabular}}
\vspace{2mm}
\caption{Description of two used multi-document summarization datasets: Multi-News and Multi-XScience.} 
\label{tab: dataset}
\end{table*}

\subsection{Data Processing}
For Multi-XScience and Multi-News datasets, the source documents are separated by a special token named ``story\_separator\_special\_tag''. The length of the input documents is restricted to 1024 tokens. In each document set, the number of tokens for one document is $\frac{1024}{N}$, where $N$ is the number of documents in a document set. 
For some shorter documents, the documents repeat themselves to fill the 1024 token quota. 
In the Multi-XScience dataset, the citations in the sources and targets are replaced by a common token `@cite'.

\subsection{Evaluation Metrics}
\label{subsec:eval_metrics}

\vspace{1mm}
\noindent \textbf{ROUGE}\footnote{The parameters of ROUGE are -c 95 -2 -1 -U -r 1000 -n 4 -w 1.2 -a -m.}
Recall-Oriented Understudy for Gisting Evaluation \cite{lin-2004-rouge} is a set of evaluation metrics for comparing the overlapping textual units between generated summaries and golden summaries, including ROUGE-1 (R-1), ROUGE-2 (R-2), ROUGE-L (R-L), ROUGE-SU (R-SU). R-1 and R-2 measure the overlapping unigrams and bigrams respectively while R-L identifies the longest co-occurring sequence of n-grams. R-SU is calculated as a statistic to measure the co-occurrence of unigram and skip-bigram. 

\vspace{1mm}
\noindent \textbf{ROUGE-WE (R-WE)}  \cite{ng-abrecht-2015-better}
is a variant of the ROUGE metric which 
replaces the hard lexical matching in ROUGE-N with a soft matching based on the cosine similarity of word embeddings. The soft matching in ROUGE-WE provides a more forgiving evaluation by not strictly requiring exact lexical matches, thus allowing for variations in word order and phrasing.

\vspace{1mm}
\noindent \textbf{BLEU}
BiLingual Evaluation Understudy \cite{Papineni-etal} introduces a brevity penalty term and computes the geometric average of the modified n-gram precision.

\vspace{1mm}
\noindent \textbf{S\textsuperscript{3}}
\cite{peyrard-etal-2017-learning} is a model-based metric that considers the features from other evaluation metrics, including R-N, R-L, R-WE and JS-divergence, to produce pyramid (pyr) and responsiveness (resp) scores. 

\vspace{1mm}
\noindent \textbf{BertScore (BS)}\footnote{The model type of BertScore is bert-base-uncased.}
\cite{Zhang2020BERTScore} measures the soft overlap of the token BERT embeddings from the machine-generated summaries and golden summaries. 

\vspace{1mm}
\noindent \textbf{Relevance (Rel)}
\cite{Simple-Theoretical} calculates cross-entropy over individually constructed probability distributions for a summary \(S\) and a source \(D\) using their own semantic units \(\omega\): $ Relevance(S,D)= \sum\limits_{\omega _{i} } \ P_{S}(\omega _{i}) \ . \ log(P_{D}(\omega _{i}))$, where probability distributions of summary and source document are given by \(P_{S}\) and \(P_{D}\) respectively.

\vspace{1mm}
\noindent \textbf{Redundancy(Red)}
\cite{Simple-Theoretical} evaluates the quality of the accumulation of information in the candidate summaries: $Redundancy(S)= \sum\limits_{\omega _{i} } \ P_{S}  (\omega _{i}) \ . \ log(P_{S}(\omega _{i}))$.

\subsection{Visualization on the Impact of Document Separators}

\begin{figure*}[]
  \centering
  \subfigure{
\begin{minipage}[t]{0.4\linewidth}
\centering
\includegraphics[width=1\textwidth]{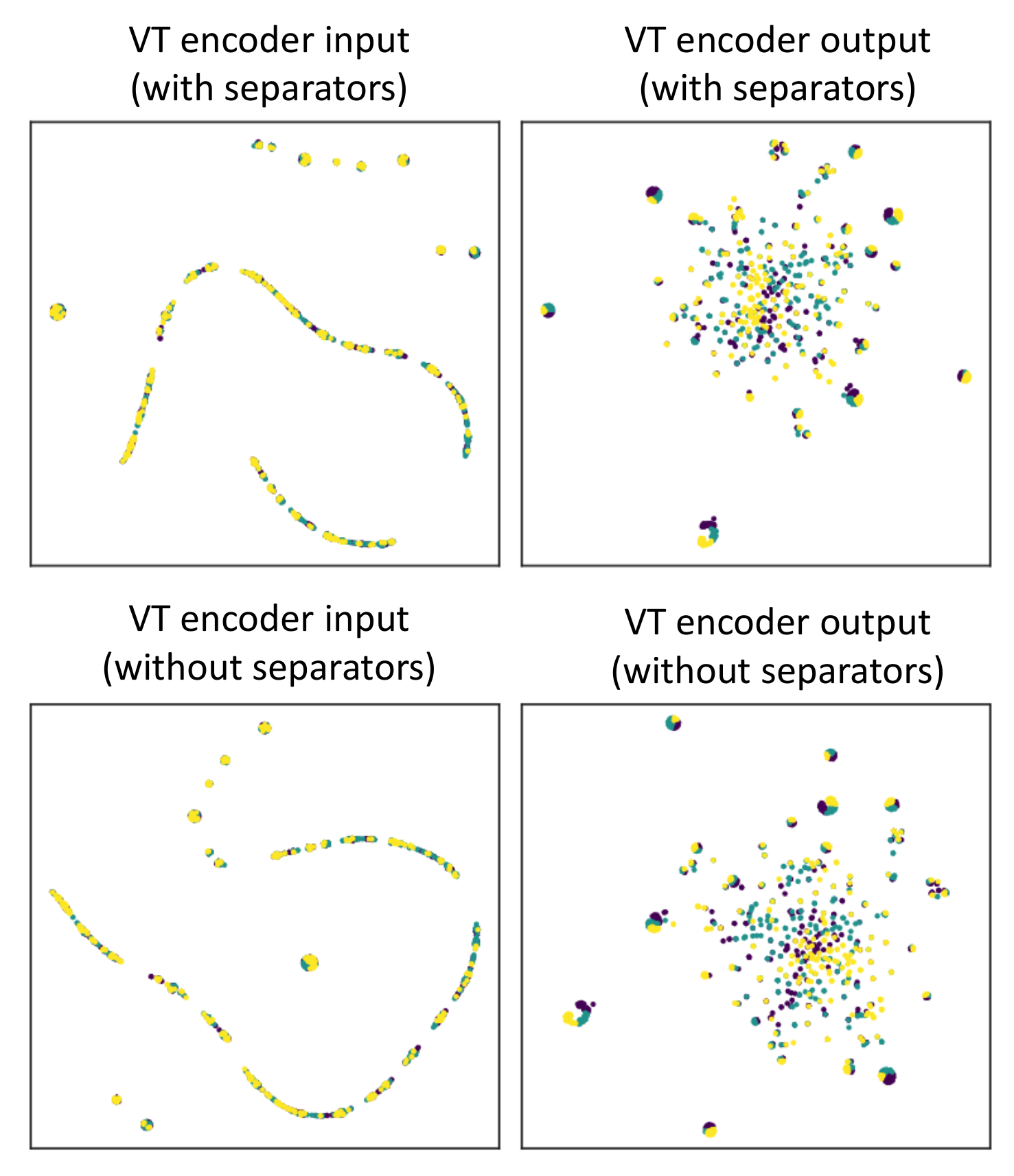}
(VT)
\label{fig:VT_sepetator}
\end{minipage}}
\subfigure{
\begin{minipage}[t]{0.4\linewidth}
\centering
\includegraphics[width=1\textwidth]{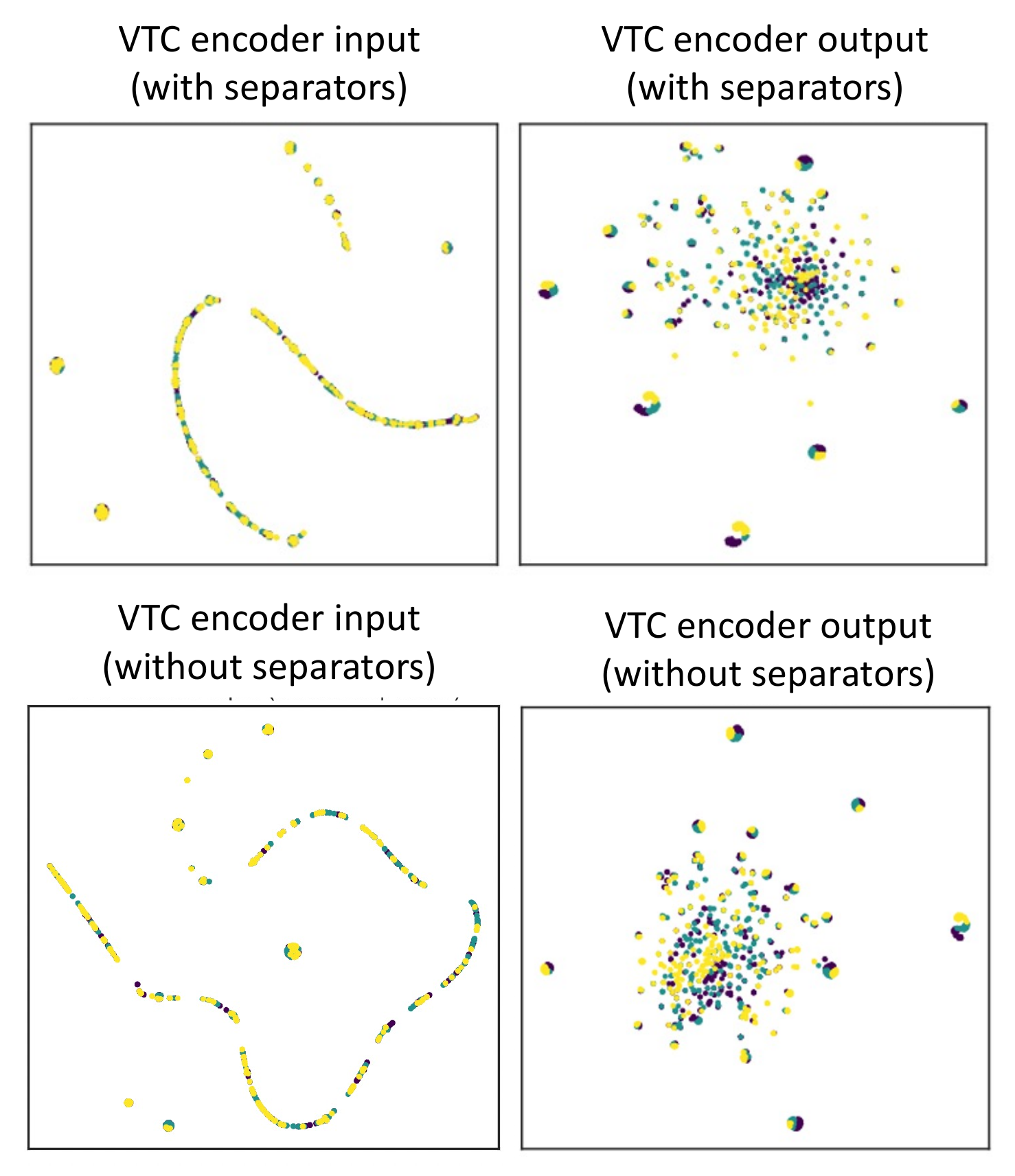}
(VTC)
\label{fig:VTC_sepetator}
\end{minipage}}
\newline
\subfigure{
\begin{minipage}[t]{0.4\linewidth}
\centering
\includegraphics[width=1\textwidth]{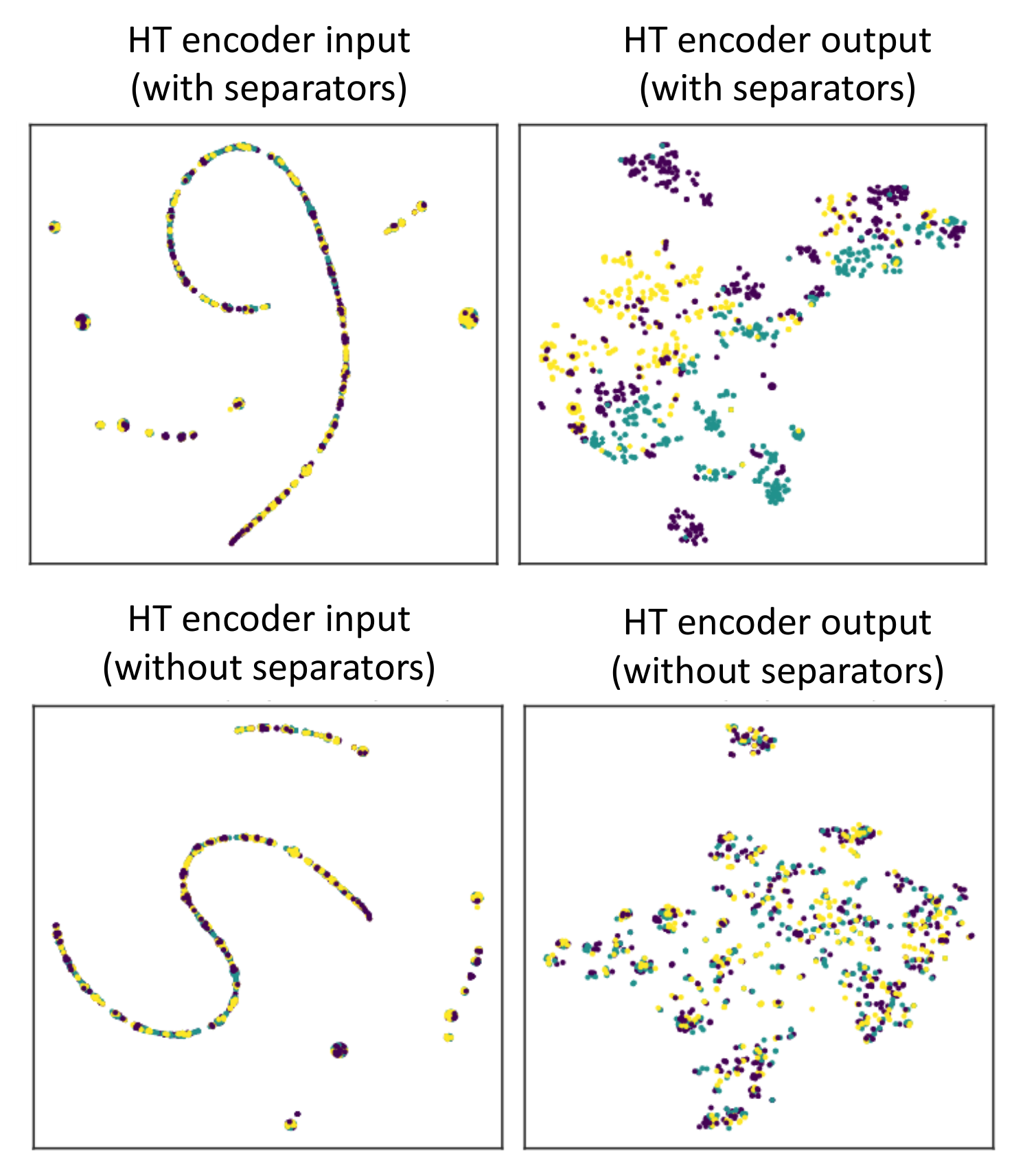}
(HT)
\label{fig:HT_sepetator}
\end{minipage}}
  \caption{t-SNE visualization of two embedding spaces on Multi-News dataset with VT, VTC and HT models: (1) token representations before feeding into the Transformer encoder; (2) token representations after feeding into the Transformer encoder. The figures in the 1\textsuperscript{st} row are the visualization with document separators and in the 2\textsuperscript{st} row are the visualization without document separators.}
\label{fig:separator}
\end{figure*}

We compare and analyze the embedding space of the tokens after they feed into the encoder with and without document separators by t-SNE visualization (Figure \ref{fig:separator}). After token representations feed into the hierarchical Transformer encoder, the cluster boundaries of documents with separators are easier to be identified in the embedding space.  Different from the hierarchical Transformer model, these two flat Transformer models have difficulties to distinguish the document cluster boundaries in the embedding space when the token representations after feed into Transformer encoder. Potentially, the hierarchical Transformer prefers more structural information of documents to compose the final summaries, while the flat Transformer does not.

\end{document}